\documentclass[letterpaper]{article} %
\usepackage[]{aaai23}  %
\usepackage{times}  %
\usepackage{helvet}  %
\usepackage{courier}  %
\usepackage[hyphens]{url}  %
\usepackage{graphicx} %
\usepackage{natbib}  %
\usepackage{caption} %
\usepackage{algorithm}
\usepackage{algorithmic}

\usepackage{newfloat}
\usepackage{listings}
\DeclareCaptionStyle{ruled}{labelfont=normalfont,labelsep=colon,strut=off} %
\floatstyle{ruled}
\newfloat{listing}{tb}{lst}{}
\floatname{listing}{Listing}
\nocopyright %

\usepackage[pagebackref=true,breaklinks=true,bookmarks=false]{hyperref}

\usepackage[capitalize]{cleveref}
\crefname{section}{Sec.}{Secs.}
\Crefname{section}{Section}{Sections}
\Crefname{table}{Table}{Tables}
\crefname{table}{Tab.}{Tabs.}

\usepackage{xcolor}
\usepackage{xspace}
\usepackage{pifont}%
\usepackage{colortbl}  %

\usepackage{caption}  %
\usepackage{subcaption}  %

\usepackage{booktabs} %
\usepackage{multirow}
\usepackage{amssymb}  %
\usepackage[toc,page]{appendix}

\definecolor{noel_color}{RGB}{10, 114, 72}
\definecolor{gray}{RGB}{153, 153, 153}
\definecolor{light_gray}{RGB}{236, 236, 236}

\newcommand{\xmark}{\ding{55}}

\newcommand{\mypara}[1]{\vspace{2pt}\noindent{\bf{#1}}}

\usepackage{enumitem}
\newlist{todolist}{itemize}{2}
\setlist[todolist]{label=$\square$}

\newcommand{\methodName}{NamedMask\xspace}

\newcommand{\mcal}[1]{\mathcal{#1}}
\newcommand{\mbb}[1]{\mathbb{#1}}

\newcommand{\gray}[1]{\textcolor{gray}{#1}}
\newcommand{\ie}{\textit{i.e.,}\xspace}
\newcommand{\eg}{\textit{e.g.,}\xspace}
\newcommand{\vhead}[1]{\rotatebox{60}{\parbox[c]{0cm}{\centering{#1}}}}

\title{NamedMask: Distilling Segmenters from Complementary Foundation Models}

\author{
    Gyungin Shin\textsuperscript{\rm 1}\hspace{7mm}
    Weidi Xie\textsuperscript{\rm 1, 2}\hspace{7mm}
    Samuel Albanie\textsuperscript{\rm 3}
}
\affiliations{
    \textsuperscript{\rm 1}Visual Geometry Group, University of Oxford, UK\\
    \textsuperscript{\rm 2}Coop. Medianet Innovation Center, Shanghai Jiao Tong University, China\\
    \textsuperscript{\rm 3}Department of Engineering, University of Cambridge, UK\\
    {\small project page: \url{https://www.robots.ox.ac.uk/~vgg/research/namedmask}}

}

\usepackage{bibentry}

\begin{document}

\maketitle

\begin{abstract}
The goal of this work is to segment and name regions of images
without access to pixel-level labels during training.
To tackle this task, we  construct segmenters by distilling the complementary strengths of two foundation models.
The first, CLIP~\cite{radford2021icml}, exhibits the ability to assign names to image content but lacks an accessible representation of object structure.
The second, DINO~\cite{caron2021iccv}, captures the spatial extent of objects but has no knowledge of object names.
Our method, termed \methodName, begins by using CLIP to construct category-specific archives of images.
These images are pseudo-labelled with a category-agnostic salient object detector bootstrapped from DINO, then refined by category-specific segmenters using the CLIP archive labels.
Thanks to the high quality of the refined masks, we show that a standard segmentation architecture trained on these archives with appropriate data augmentation achieves impressive semantic segmentation abilities for both single-object and multi-object images.
As a result, our proposed \methodName performs favourably against a range of prior work on five benchmarks including the VOC2012, COCO and large-scale ImageNet-S datasets.

\end{abstract}

\section{Introduction}

Semantic segmentation is a task that entails grouping and naming coherent regions of images.
It has a broad range of applications spanning domains such as autonomous driving, manufacturing and medicine.
A key barrier to automating this task through supervised learning is the requirement for pixel-level segmentation annotations, 
which can be extremely costly to obtain (e.g. 1.5 hours per image when accounting for quality control~\cite{cordts2016cvpr}).

The emerging paradigm of \textit{foundation models}~(models that have been pre-trained on broad data and can be adapted to a wide range of downstream tasks) has yielded striking gains for many machine perception problem domains~\cite{bommasani2021opportunities}.
There is therefore considerable interest in determining whether such models can alleviate the prohibitive annotation costs associated with semantic segmentation. In this vein, MaskCLIP~\cite{zhou2022maskclip} demonstrated the potential benefits of leveraging the representation learned by CLIP~\cite{radford2021icml} to perform ``annotation-free''\footnote{This setting can be equivalently referred to as \textit{zero-shot transfer} in the terminology of~\citet{radford2021icml}.} segmentation with no prior knowledge of the target domain.
However, unless segmentation masks are available from at least some of the categories to guide pseudo-labelling, 
segmentation quality remains far from supervised performance.
ReCo~\cite{shin2022reco} considered an alternative formulation in which only the \textit{names} of the target categories (but no images from the target domain) are available during training.
By coupling the retrieval capabilities of CLIP to curate archives of images belonging to specific categories with a co-segmentation algorithm, ReCo obtained improvements over MaskCLIP 
but still struggles to produce precise object segmentations.

\begin{figure}
    \centering
    \includegraphics[width=.4725\textwidth]{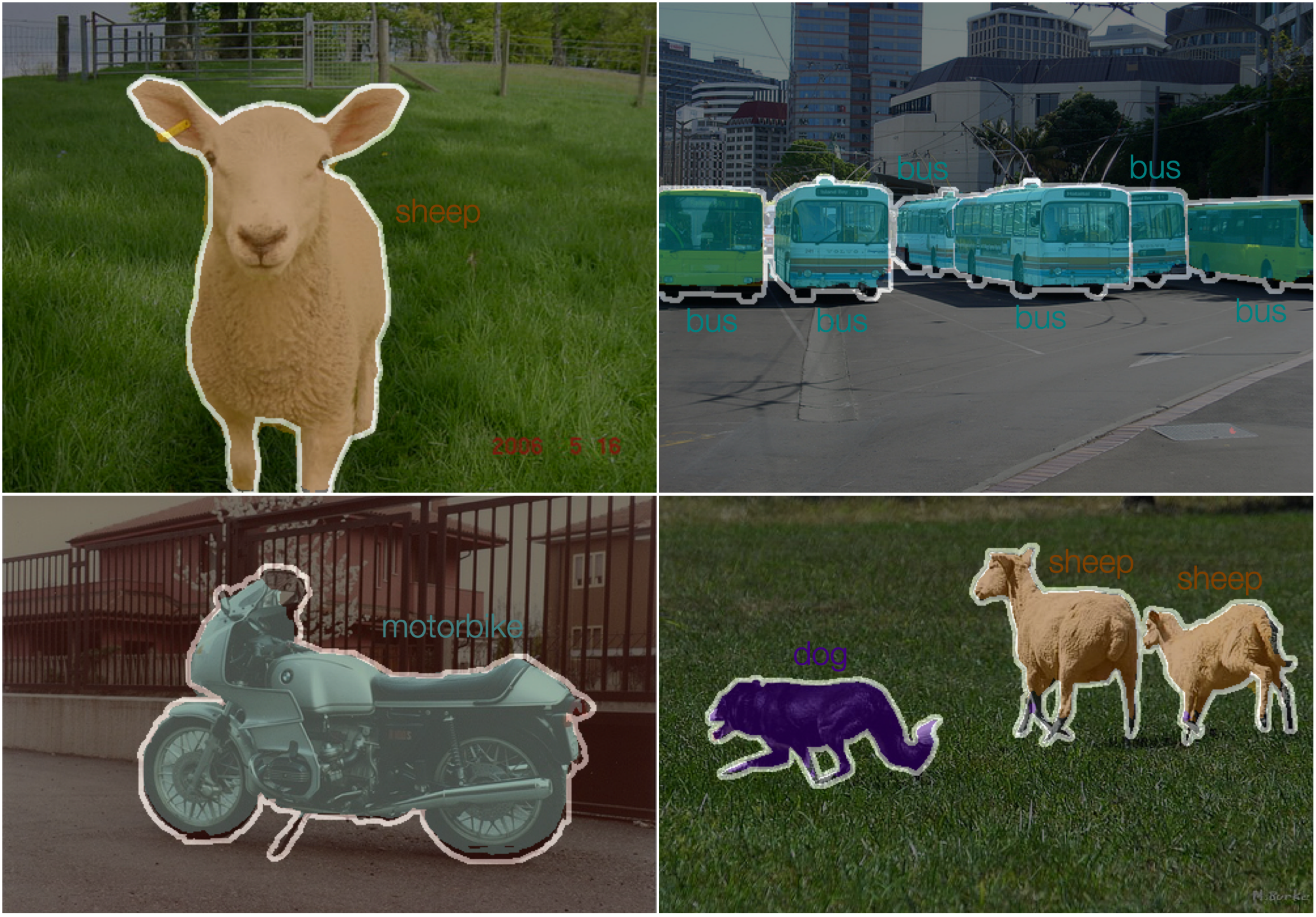}
    \caption{We propose \methodName, a segmenter distilled from the complementary capabilities of CLIP and DINO.
    \textit{With no access to pixel-level annotation}, \methodName not only accurately segments single objects (left) but also multiple objects (right) within an image. 
    White pixels denote ignored regions.
    The images are drawn from the VOC2012 benchmark.}
    \vspace{-5mm}
    \label{fig:teaser}
\end{figure}

In this work, we build upon the ReCo framework and revisit the mechanism through which it obtains pseudo-masks with semantic labels. 
Drawing inspiration from recent work showing that DINO features can be employed to perform unsupervised salient object detection~\cite{shin2022unsupervised,melas2022deep,wang2022self}, 
our first step is to replace the fragile co-segmentation of ReCo with a more robust category-agnostic object segmentation facilitated by DINO.
We then exploit the naming capabilities of CLIP to assign the category label from each archive of images to these segmentations to enable the training of category-specific ``expert'' segmenters that refine the quality of the archive segmentations.
Finally, we train a single semantic segmentation model on the resulting collection that is capable of segmenting objects from any category that is represented in the archives, using copy-paste augmentation~\cite{ghiasi2021cvpr} to improve generalisation to images of multiple objects.
We show that our approach, which we term \methodName, achieves substantial gains in performance for semantic segmentation of objects (see~\cref{fig:teaser} for examples).

Our contributions are:
(1) We propose \methodName, a framework for segmenting and naming objects without pixel-level annotation by distilling the complementary strengths of CLIP and DINO;
(2) We provide extensive experiments to demonstrate the improvements brought by \methodName over prior semantic segmentation approaches that also make use of language-image pretraining.

\section{Related work}

Our approach relates to prior work on
\textit{unsupervised semantic segmentation},
\textit{semantic segmentation with language-image pretraining}
and \textit{salient object detection}.
We discuss connections to each of these next.

\mypara{Unsupervised semantic segmentation.}
By coupling deep neural networks with creative learning objectives, substantial progress has been made towards unsupervised semantic segmentation.
Examples of learning signals that have been constructed without labels include
expectation-maximisation over segments~\cite{hwang2019segsort},
mutual information maximisation~\citep{ji2019iccv,ouali2020eccv},
contrasting proposals~\cite{gansbeke2022maskdistill},
complementary signals from LiDAR and vision~\cite{vobecky2022arxiv} and
feature correspondence distillation~\cite{hamilton2022iclr}.
In contrast to name-only segmentation,
these methods do not make use of language-image pretraining or require access to the target category list during training.
They do, however, require the use of a small number of images labelled with segments (typically the test set itself) together with the Hungarian algorithm to assign names to segment predictions, 
or otherwise employ nearest-neighbour lookup on a held-out set of images with segmentation masks.

\mypara{Annotation-free semantic segmentation using language-image model.}
Several recent works have sought to leverage the zero-shot transfer capabilities of CLIP~\citep{radford2021icml} to perform semantic segmentation with no access to paired data (images labelled with categories or segments) from the target domain.
MaskCLIP~\cite{zhou2022maskclip} illustrated the potential of using CLIP for semantic segmentation in a zero-shot transfer setting (a setting that they term ``annotation-free'').
A recent example of such line of research is ReCo~\citep{shin2022reco}, which curates unlabelled images into examples of concepts with CLIP, then applies a co-segmentation algorithm to derive semantic segmentation training data.
While ReCo achieves promising results, it fails to coherently pseudo-label objects and thus (as we show through experiments) does not lead to high-quality object segmentations.
In this work, we compare directly with ReCo and demonstrate the substantial gains in performance that can be attained by bootstrapping the category-agnostic pseudo-labels enabled by DINO.

\mypara{Unsupervised salient object detection.}
A range of work has sought to perform salient object detection (the task of segmenting foreground objects) without human annotation~\cite{Zhang_2017_ICCV,bielski2019emergence,voynov2020unsupervised}.
One notable trend amongst recent approaches has been the application of spectral clustering in combination with self-supervised features~\cite{melas2022deep,wang2022self,shin2022unsupervised}.
In this work, we build on the SelfMask approach of~\cite{shin2022unsupervised} to provide a robust category-agnostic segmenter for \methodName.

\begin{figure*}[t]
    \centering
    \includegraphics[width=\textwidth]{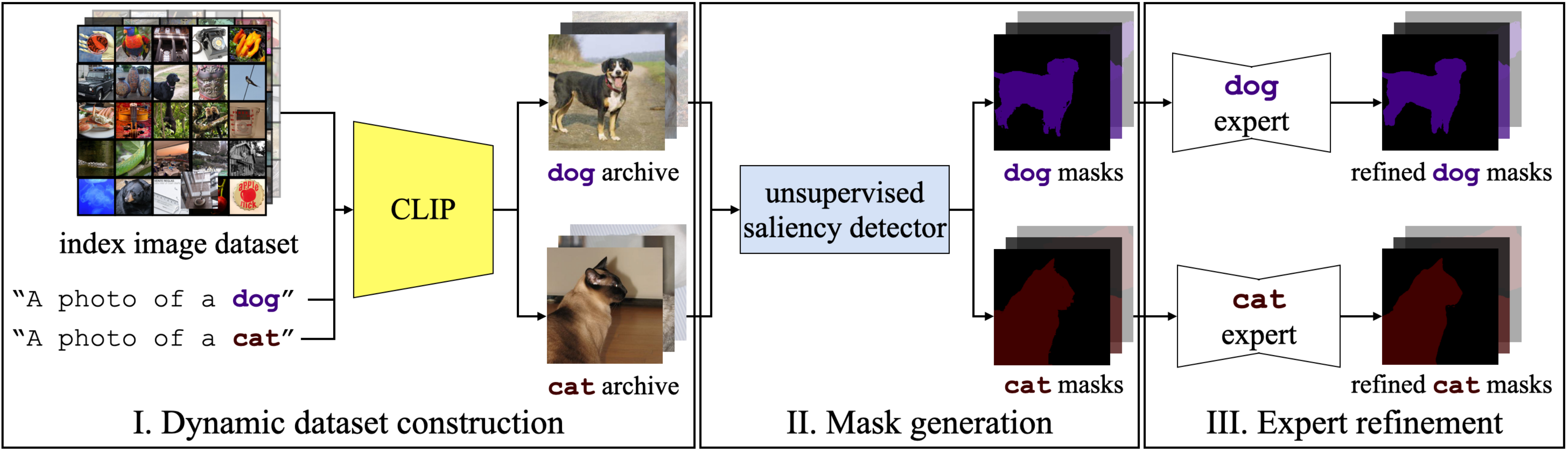}
    \caption{Overview of the proposed pipeline used to construct the \methodName training dataset for semantic segmentation. 
    Given an image archive for a concept retrieved by CLIP (left), we generate masks using an unsupervised saliency detector (middle). 
    We refine the segmentations of each category by a class expert trained with the constructed image-mask pairs (right). Using the retrieved images and their refined segments, we train \methodName to generate a segmenter capable of predicting a set of pre-defined categories (omitted in the figure for simplicity).}
    \vspace{-5mm}
    \label{fig:overview}
\end{figure*}
\section{Method}~\label{method}
In this section, 
we formulate the semantic segmentation task considered in this work~(\cref{method:task}) and the method, \methodName, 
that we propose to tackle it~(\cref{method:languagemask}).

\subsection{Task formulation and terminology}~\label{method:task}
Our objective is to perform \textit{semantic segmentation}: 
for a given image, $x \in \mathbb{R}^{3 \times H \times W}$, we aim to assign a label, $c$, from among a finite set of categories, $\mathcal{C}$, to each pixel location $\omega \in \{1, \dots, H\} \times \{1, \dots, W\}$ of $x$.
To facilitate cost-effective scaling, 
we aim to do so without access to any form of pixel-level annotation. 
To this end, we propose to exploit the perceptual grouping of objects 
and their semantic categorisation offered by two foundation models.
Specifically, we leverage the semantic categorisation capabilities of CLIP derived through large-scale language-image pretraining and the perceptual grouping capabilities of DINO derived from vision-only pretraining.

\vspace{2pt}
\noindent \textbf{Terminology}. 
To date, a wide array of methods have been proposed to tackle the problem of semantic segmentation with different levels of supervision (\textit{fully unsupervised}, 
\textit{unsupervised but with supervised pretraining}, \textit{weakly-supervised} etc.).
However, the terminologies used to describe these levels of supervision are not always clear or consistent.
We therefore first aim to clarify the annotation regime in which we operate and how it is closely related to prior work.

In particular, we consider a setting that 
we term \textit{Segmentation Leveraging Only Weak Pretraining}~(SLOWP).
SLOWP methods make no use of pixel-level annotation and are characterised by pretraining on data that is either:
(1) \textit{weakly-labelled} (e.g. with class labels or alt-text) and does not derive from the target distribution; or
(2) \textit{unlabelled} and may or may not derive from the target distribution.
Within the space of SLOWP methods, we further distinguish three sub-categories that more precisely characterise the knowledge that the method possesses about the segmentation task used to evaluate the model:
(i) \textit{Zero-shot transfer} assumes no knowledge of the target distribution (images or category names) during training;
(ii) \textit{Name-only transfer} assumes access (during training) to the list of category names that are to be used for the target segmentation task, but does not assume access to any images from the target distribution;
(iii) \textit{Name-and-image transfer} assumes access (during training) to the list of category names in the target segmentation task \textit{and} access to unlabelled images from the target distribution.

To relate these categories to prior work, note that MaskCLIP~\cite{zhou2022maskclip} represents a \textit{SLOWP zero-shot transfer} method: it uses language-image pretraining (via CLIP) and does not make use of the target categories during training.
ReCo~\cite{shin2022reco} typically represents a \textit{SLOWP name-only transfer} method: it uses language-image pretraining (via CLIP) and image classification pretraining (via DeiT-S/SIN~\cite{naseer2021intriguing}) and has access to target category names for constructing classifiers.

In this work, we propose a method, \methodName, that operates effectively in both the \textit{SLOWP name-only transfer} and \textit{SLOWP name-and-image transfer} scenarios.
We describe \methodName next.

\subsection{\methodName}~\label{method:languagemask}
\methodName is trained in a sequence of four stages:
(1) For a given list of target categories, we perform dynamic dataset construction by curating archives of images for each category from an unlabelled image collection using CLIP;
(2) For each image in each archive, we predict a category-agnostic object mask with an unsupervised saliency detector;
(3) We refine the predicted masks with a category-specific ``expert'' segmenter, which is self-trained with the generated image-mask pairs within each archive;
(4) We distill a segmenter using the image archives and their refined masks as pseudo-labels.
An overview of the first three stages is provided in~\cref{fig:overview}, and each stage is detailed in the following.

\mypara{Dynamic archive construction.}
To create a data set containing images of categories of interest, we follow the approach proposed by ReCo~\cite{shin2022reco} and curate an archive of images for each concept using an image-language model (\ie CLIP). Formally, given an image encoder $\phi_{\mcal{I}}$ and a text encoder $\phi_{\mcal{T}}$ of CLIP, 
we curate one archive from an unlabelled image collection $\mcal{U}$ for each category $c$ of interest.
We do so by selecting the $n$ images among the collection whose visual embeddings $\phi_{\mcal{I}}(x_{i}) \in \mbb{R}^e$ lie closest to the text embedding\footnote{Details of the prompt used to construct the text embedding can be found in the supplementary material.} $\phi_{\mcal{T}}(c) \in \mbb{R}^e$ of $c$. 
That is, 
\begin{equation}
    \mcal{A}_c = \{x_i~|~i \in \mbox{arg top}k~ [ \phi_{\mcal{I}}(\mcal{U}) \cdot \phi_{\mcal{T}}(c) ] \},
\end{equation}
where $\mcal{A}_c$ denotes the image archive for category $c$ and arg top$k$ returns the indices of its arguments with the $k$ largest values.
In this way, we dynamically construct a data set comprising a collection of $|\mathcal{C}|$ archives (one for each category).

\mypara{Mask generation.}
To produce category-agnostic object segmentations for the images within each archive, we adopt the SelfMask~\cite{shin2022unsupervised} unsupervised salient object detection method. 
SelfMask learns to perform salient object detection by first performing spectral clustering on DINO features across unlabelled images, 
then using these clusters as pseudo-labels to train a variant of MaskFormer segmenter~\cite{cheng2021per}.
Given the SelfMask saliency detector $\psi_s$, we first predict a category-agnostic saliency map $y_i = \psi_s(x_i)$ $\in \{0, 1\}^{H \times W}$ for each image $x_i \in \mbb{R}^{3 \times H \times W}$ in each archive.
We then simply assign to each category-agnostic saliency map the category label $c$ of the archive that contains the image.
This produces a collection of images annotated with saliency masks and corresponding category labels.

\mypara{Mask refinement through category experts.}
The category-agnostic saliency detector employed in the previous stage is unaware of the category of objects that it is being used to segment. 
We hypothesise that a segmenter will produce superior segmentations when it is given knowledge of the specific category of objects that it is required to segment, and thus will produce improved pseudo-masks for training \methodName. 
To instantiate this idea, we refine the category-agnostic predictions made by the saliency detector with a segmenter $\psi_c$, which specialises in segmenting regions corresponding to category $c$.
To this end, we train a segmenter $\psi_c$ to assign regions to either the category $c$ or the background class for each image in $\mcal{A}_c$,
as a pixel-level one-vs-all binary classification task.
We then use the predictions obtained by $\psi_c$ as pseudo-masks for category $c$. 
We show through experiments in~\cref{experiments:ablation} that this simple approach yields superior segmentation training data relative to using SelfMask predictions directly.

Note that for cases when there are a large number of target categories (\eg 919 categories in ImageNet-S~\cite{gao2021arxiv}),
training one expert per class can be computationally expensive.
For such cases, we group the relevant categories by applying $k$-means clustering to the text embeddings of the categories extracted from a CLIP text encoder and train an expert for each category group.

\mypara{Training \methodName.}
Given the resulting collection of image archives annotated by category-specific segmenters, 
\methodName is produced by simply training a standard semantic segmentation architecture using a cross-entropy loss.
Thus, self-training produces a segmenter that exploits the complementary information encoded by two different foundation models,
where the visual-only model~(\ie DINO) implicitly captures the perceptual grouping of objects, and the ability to name categories derives from the language-image model~(\ie CLIP).

\section{Experiments}

In this section, we begin by describing the datasets considered for our experiments, implementation details, and our ablation study. 
We then compare our model to state-of-the-art unsupervised semantic segmentation~(USS) methods and approaches that leverage only weak pretraining~(SLOWP).

\subsection{Datasets}

\mypara{Evaluation benchmarks.}
We consider five segmentation benchmarks including COCO~\cite{lin2014microsoft}, CoCA~\cite{zhang2020gicd}, Cityscapes~\cite{cordts2016cvpr}, PASCAL VOC2012~\cite{Mark2015VOC}, and ImageNet-S~\cite{gao2021arxiv}.
COCO consists of 118,287 and 5,000 images for train and validation splits with 80 object categories and a background class and CoCA comprises 1,295 images of 80 object categories with a background.
Cityscapes contains 2,975 and 500 urban scene images for training and validation splits with 30 categories among which we pick 14 object categories to evaluate based on the original paper~\cite{cordts2016cvpr}.
VOC2012 is composed of 1,464 training and 1,449 validation images with 21 categories including background, and the large-scale ImageNet-S~\cite{gao2021arxiv} dataset comprises 9,190 train, 12,419 validation, and 27,423 test images with precise pixel-level annotations. There are three variations of ImageNet-S: ImageNet-S$_{50}$, ImageNet-S$_{300}$, and ImageNet-S$_{919}$, 
consisting of 50, 300, and 919 semantic categories of ImageNet1K~\cite{deng2009imagenet}, respectively.

We use the VOC2012 train split and the ImageNet-S$_{300}$ validation split 
for our ablation studies, and compare our models to previous USS and SLOWP methods on CoCA, the validation split of COCO, Cityscapes, and VOC2012, and the test split of ImageNet-S.

\setlength{\tabcolsep}{3pt}
\begin{table*}[t]
    \small
    \centering
    \begin{tabular}{c ccccc ccccc ccccc ccccc c}
    model &
    \multirow{1}{*}{\vhead{aeroplane}}&
    \multirow{1}{*}{\vhead{bicycle}}&
    \multirow{1}{*}{\vhead{bird}}&
    \multirow{1}{*}{\vhead{boat}}&
    \multirow{1}{*}{\vhead{bottle}}&
    \multirow{1}{*}{\vhead{bus}}&
    \multirow{1}{*}{\vhead{car}}&
    \multirow{1}{*}{\vhead{cat}}&
    \multirow{1}{*}{\vhead{chair}}&
    \multirow{1}{*}{\vhead{cow}}&
    \multirow{1}{*}{\vhead{dining-table}}&
    \multirow{1}{*}{\vhead{dog}}&
    \multirow{1}{*}{\vhead{horse}}&
    \multirow{1}{*}{\vhead{motorbike}}&
    \multirow{1}{*}{\vhead{person}}&
    \multirow{1}{*}{\vhead{potted~plant}}&
    \multirow{1}{*}{\vhead{sheep}}&
    \multirow{1}{*}{\vhead{sofa}}&
    \multirow{1}{*}{\vhead{train}}&
    \multirow{1}{*}{\vhead{tv/monitor}}&
    \multirow{1}{*}{avg.}\\
    \midrule
    SelfMask
    &78.5
    &26.8
    &68.0
    &59.2
    &22.9
    &69.3
    &43.1
    &80.7
    &14.0
    &70.2
    &20.3
    &74.2
    &70.1
    &57.7
    &41.5
    &20.6
    &74.3
    &18.4
    &64.7
    &25.0
    &50.0\\
    experts
    &80.2
    &29.0
    &72.9
    &65.0
    &30.4
    &74.9
    &48.9
    &82.4
    &15.8
    &77.8
    &27.2
    &75.2
    &74.5
    &62.1
    &43.1
    &21.3
    &74.9
    &26.7
    &69.8
    &36.4
    &54.4\\
    \bottomrule
    \end{tabular}
\caption{\textbf{Category experts produce better quality segmentation masks than the baseline unsupervised saliency detector.} We report segmentation performance for each method on Pascal VOC2012. The performance metric is (class-wise) IoU (\%).}
\vspace{-5mm}
\label{tab:specialists}
\end{table*}

\mypara{Image collections.}
To curate image archives for each category, we use two unlabelled image collections:
(1) For experiments on PASCAL VOC2012, we use the ImageNet1K training set without labels, following~\cite{shin2022reco}.
(2) For experiments on ImageNet-S, we use unlabelled images from LAION-5B~\cite{schuhmann2022web}. 
For the latter, we implement the archive curation process using the CLIP feature index provided with the LAION-5B release\footnote{\url{https://laion.ai}}. 
Since the LAION-5B dataset was collected with limited manual curation, %
we apply a face detector to all images and discard any image containing a visible human face~\cite{Deng2020CVPR}.
We refer the reader to the supplementary material for further details about the usage of the LAION-5B dataset.

\subsection{Implementation details}
We conduct the experiments on a single A100 NVIDIA graphic card with PyTorch~\cite{adam2019neurips}. 
Code will be made publicly available.

\mypara{Network architecture and optimisation.}
We use DeepLabv3+~\cite{chen2018eccv} with a ResNet50~\cite{he2016cvpr} backbone for both category experts and \methodName. 
We initialise the backbone with DINO~\cite{caron2021iccv} that is pretrained on ImageNet~\cite{russakovsky2015imagenet} in a self-supervised manner.
For expert training, we adopt a lightweight learning schedule of 5K gradient updates with a batch size of 8 for COCO, CoCA, Cityscapes, and VOC2012 and 10K updates with a batch size of 16 for ImageNet-S.
For \methodName, we train the model with 20K iterations for COCO, CoCA, Cityscapes, and VOC2012 and 80K iterations for ImageNet-S. 
We use standard data augmentations such as random scaling, random cropping and colour jittering. 
We use Adam optimiser with an initial learning rate of 0.0005 and a weight decay of 0.0002. We decay the learning rate with the Poly learning rate~\cite{chen2018eccv}.

To curate category archives from ImageNet and LAION-5B, the ViT-L/14@336px and ViT-L/14 variants of CLIP are employed repectively. %
For our unsupervised saliency detection method, 
we adopt the model from SelfMask~\cite{shin2022unsupervised}, 
and apply a bilateral solver~\cite{barron2016fast} to predictions from SelfMask as a post-processing step.

\mypara{Inference.}
When evaluating on ImageNet-S, 
images are resized such that their larger dimension is $1024$ pixels while preserving their aspect ratio. 
For evaluation, the predictions of the model are then resized back to the original resolution to match the ground-truth mask by using a bilinear upsampler. 
For the ImageNet-S$_{300}$ validation set (used in our ablation study), we resize the shorter side of images to 384 with a maximum length for the larger dimension of 512 pixels.
For the other benchmarks, we use the original resolution of the images.

\mypara{Metric.}
Following the common practice, we employ intersection-over-union (IoU) to measure a class-agnostic mask quality and mean IoU (mIoU) to evaluate the performance of semantic segmentation.

\subsection{Ablation study}\label{experiments:ablation}
\begin{figure}[!t]
    \centering
    \includegraphics[width=.48\textwidth]{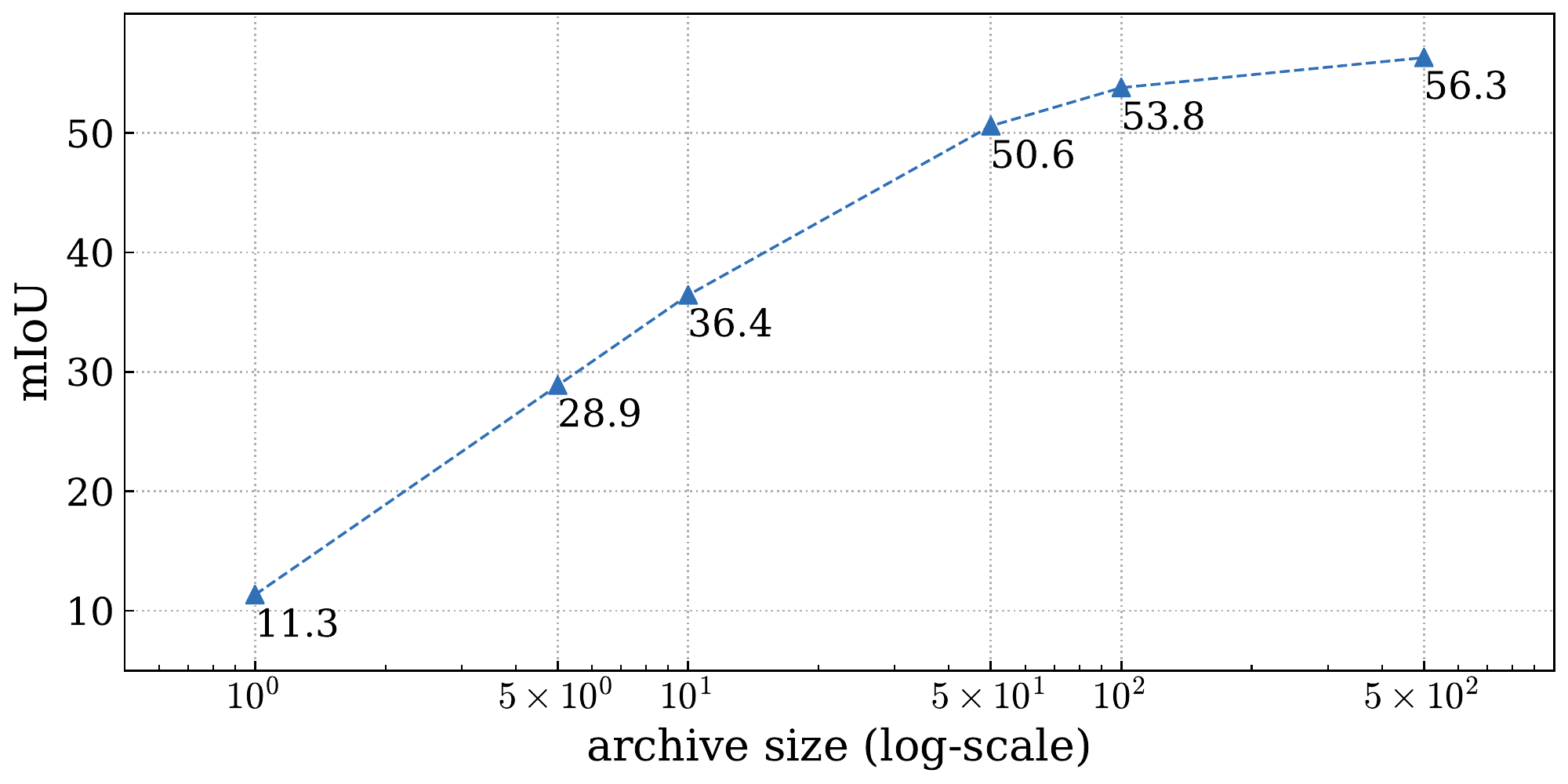}
    \caption{\textbf{Larger image archives produce better segmenters.}
    Here, ``archive size'' denotes the number of images retrieved by CLIP to curate an image archive for each category.}
    \vspace{-5mm}
    \label{fig:archive-size}
\end{figure}

In this section, we present a thorough ablation study on each component of our proposed \methodName, namely, the influence of archive size, the influence of adopting category experts and the effect of the number of category experts. 
We also investigate the influence of adopting copy-paste augmentation for segmenting images with multiple objects.

\mypara{Effect of archive size.} 
Unlike supervised approaches for which it is costly to acquire annotations, the dataset creation process for \methodName can be easily scaled. 
To investigate the influence of the number of images used for training, we vary the size of archive curated by CLIP for each category, from 1 to 500 images, 
and train \methodName on the resulting images with corresponding pseudo-labels obtained from SelfMask. 
As for quantitative evaluation, we adopt the VOC2012 training set and report numbers in Fig.~\ref{fig:archive-size}.
Note that, the training is done only on the constructed archive of ImageNet images,  with pseudo labels acquired from SelfMask (\ie no category experts have been introduced at this stage).

As shown in Fig.~\ref{fig:archive-size}, that the archive size plays an important role in the performance of our model, 
monotonically increasing with the number of images for an archive. 
For the remaining experiments, we curate 500 images per archive.

\mypara{Effect of category experts on mask quality.} 
As described in~\cref{method:languagemask}, 
we propose to refine the pseudo-labels from SelfMask with category-specific experts, 
which are trained to distinguish foreground and background pixels.

\begin{figure*}[t]
    \centering
    \includegraphics[width=.98\textwidth]{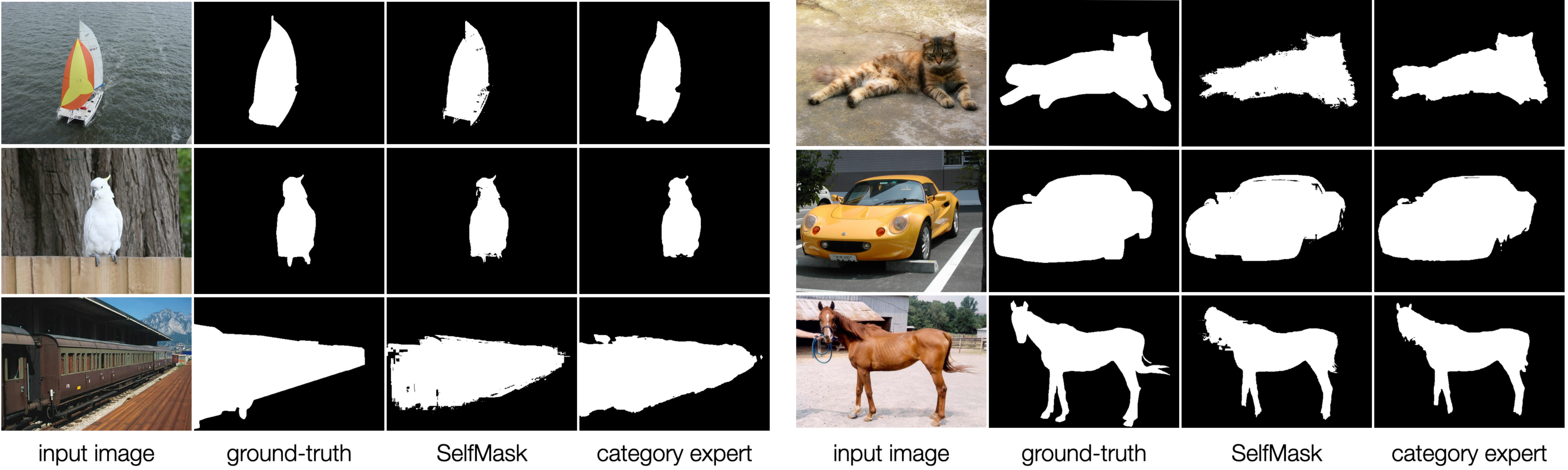}
    \caption{\textbf{Category experts refine the masks provided by an unsupervised saliency detector (\ie SelfMask).} The images are selected from VOC2012. 
    Zoom in for details.
    }
    \vspace{-5mm}
    \label{fig:mask-comparison}
\end{figure*}

To compare the quality of category-agnostic masks generated by SelfMask and class-specific masks by an expert,  
we evaluate compare their predictions on 20 object categories from the VOC2012 train split.
Specifically, we train 20 category experts on image archives constructed by retrieving from ImageNet1K. 
In~\cref{tab:specialists}, 
we show the (class) IoU of each category.
We observe that the experts consistently outperform SelfMask across all categories. 
For a qualitative comparison, 
we visualise the examples predictions from both SelfMask and category experts in~\cref{fig:mask-comparison}.
In the following experiments, we produce pseudo-masks from an expert for each category by default.

\begin{table}[!t]
\centering
\begin{tabular}{@{}lcc@{}}
\toprule
model                              & \# experts & avg. IoU \\ \midrule
SelfMask & - & 62.7 \\
                                   & 1          & 63.3     \\
                                   & 30         & 64.1     \\
                                   & 60         & 64.0     \\
\multirow{-4}{*}{category experts} & 90         & 63.9     \\ \bottomrule
\end{tabular}
\caption{Effect of grouping semantically relevant categories for category expert training on the ImageNet-S$_{300}$ validation split.}
\label{tab:category-group}
\end{table}

\begin{table}[t]
\centering
\begin{tabular}{@{}lcccc@{}}
\toprule
model & copy-paste & single obj. & multi-obj. & all\\ \midrule
\gray{SelfMask + CLIP}& \gray{-}      & \gray{63.3} & \gray{42.1} & \gray{50.4} \\
Ours     & \xmark     & 67.0          & 50.5          & 56.6 \\
Ours     & \checkmark & \textbf{68.0} & \textbf{53.6} & \textbf{58.7} \\ \bottomrule
\end{tabular}
\caption{
\textbf{Copy-paste augmentation helps the model to segment multiple objects in an image.} The performance is measured in mIoU (\%). A baseline model is marked in \gray{gray}. Best score for each column is highlighted in \textbf{bold}. 
}
\vspace{-5mm}
\label{tab:copy-paste}
\end{table}
\mypara{Training an expert for a category group.}
When there are numerous classes that are semantically close to one another, training individual category experts may become prohibitively expensive.
We therefore group categories by applying $k$-means to text embeddings of the categories into $k$ concept groups as described in Sec.~\ref{method:languagemask}.

To show the effect of $k$, we evaluate the category experts, 
by varying the number of category groups on the ImageNet-S$_{300}$ validation set. In Tab.~\ref{tab:category-group}, we report average IoU over 300 categories for $k$=\{30, 60, 90\} which corresponds to 10\%, 20\%, and 30\% of the total number of classes, respectively.
As baselines, we also show average IoU for SelfMask and a single expert for 300 categories (\ie $k$=1).
As can be seen, training an expert always shows higher score than the baseline, even when $k$ is set to 1. However, when we set the number of experts higher than 30, the performance appears to saturate.
We conjecture that this is because there is a trade-off between the number of different images which an expert can see during training and average semantic relevance within a category group. That is, when $k$ is large, the average number of categories assigned to one group tends to be low, which decreases the total number of images used for training an expert for the group. In contrast, when $k$ is small, the total number of images for an expert become large at the cost of reduced semantic relevance in a class group.

For this reason, we group categories of ImageNet-S$_{300}$ and -S$_{919}$ into 30 and 90 which (approx.) accounts for 10\% of total classes, respectively.

\mypara{Effect of copy-paste on segmenting multiple objects.}
In contrast to the salient object detectors and category experts which segment an object of a single category or a group of similar categories within an image, our model can be readily trained to segment multiple objects of different categories by employing the copy-paste augmentation~\cite{ghiasi2021cvpr}.
To demonstrate this, we evaluate two \methodName models, 
trained with and without copy-paste augmentation on the images containing a single object or multi-objects from possibly different categories. 
As a baseline, we also evaluate segmentation of SelfMask whose semantic label is decided by applying CLIP to a given image.
As shown in Tab.~\ref{tab:copy-paste}, when validated on the VOC2012 training split, copy-paste brings a notable gain in performance by 4.1 mIoU compared to the model trained without copy-paste.
We therefore adopt the copy-paste augmentation as a default setting in the remaining experiments.

\begin{table}[!t]
\centering
\setlength\tabcolsep{2pt}
\begin{tabular}{@{}lc@{}}
\toprule
model & dataset \\ \midrule
MaskCLIP & WebImageText \\
ReCo & WebImageText, (Stylized-)ImageNet\\
\methodName & WebImageText, ImageNet\\ \bottomrule
\end{tabular}
\caption{Datasets employed in training each model.} ReCo utilises both Stylized-ImageNet and ImageNet.
\vspace{-5mm}
\label{tab:dataset-info}
\end{table}
\begin{table}[!t]
\centering
\setlength\tabcolsep{2pt}
\begin{tabular}{@{}lc ccc@{}}
\toprule
model & transfer type & COCO & CoCA & Cityscapes$_{obj}$\\ \midrule
MaskCLIP & zero-shot  & 5.3 & 3.1 & 6.1 \\
ReCo$^\dagger$ & name-only & 17.1 & 16.9 & 14.1 \\
\methodName & name-only & \textbf{27.7} & \textbf{27.3}& \textbf{18.2}\\ \bottomrule
\end{tabular}
\caption{
Comparison to previous segmentation leveraging only weak pre-training (SLOWP) methods on the COCO, CoCA, and Cityscapes$_{obj}$ benchmarks in terms of mIoU.
Highest scores on each benchmark are in \textbf{bold}. $^\dagger$initialises the backbone with Stylized-ImageNet pre-training.}
\vspace{-3mm}
\label{tab:comp-slowp}
\end{table}

\begin{table}[!htb]
\centering
\setlength\tabcolsep{3pt}
\begin{tabular}{@{}lclc@{}}
\toprule
model & transfer type & backbone & mIoU \\ \midrule
\textit{USS}&&\\
Inst. Disc.  & -  & ResNet50 & 4.3 \\
MoCo & - & ResNet50 & 3.7 \\
InfoMin & - & ResNet50 & 4.4 \\
SwAV & - & ResNet50 & 4.4 \\
MaskContrast & - & ResNet50$^\dagger$ & 35.0 \\
MaskDistill & - & ResNet50$^\dagger$ & \textbf{45.8} \\ \midrule
\textit{SLOWP}\\
MaskCLIP$^*$ & zero-shot & ResNet50 &  29.1\\
ReCo$^*$$^\ddagger$ & name-only & DeiT-S/16 & 34.2\\
\methodName & name-only & ResNet50 & \textbf{59.2}\\ \bottomrule
\end{tabular}
\caption{
Comparison to existing unsupervised semantic segmentation (USS) and segmentation leveraging only weak pre-training (SLOWP) methods on the PASCAL VOC2012 validation set.
Numbers for \textit{USS} methods are from MaskDistill.
$^*$Re-implemented and adapted by us to predict a background class.
$^\dagger$uses dilated ResNet~\cite{yu2016iclr}. $^\ddagger$initialises the backbone with Stylized-ImageNet~\cite{geirhos2018iclr} pre-training.
Highest scores of each kind of methods are in \textbf{bold}.}
\vspace{-5mm}
\label{tab:comp-voc}
\end{table}

\subsection{Comparison to state-of-the-art methods}
To describe the effectiveness of our approach, we compare  \methodName against existing approaches that fall into the proposed segmentation leveraging only weak pretraining (SLOWP) setting.
Specifically, we consider and re-implement MaskCLIP~\cite{zhou2022maskclip} with the \textit{zero-shot transfer} setting (\ie the annotation-free setting in their paper) and ReCo~\cite{shin2022reco} with the \textit{name-only transfer} or and \textit{name-and-image transfer} setting.
As described in Sec.~\ref{method:task}, the transfer type of each SLOWP method is determined by whether it has access to either category names or unlabelled images from the evaluation benchmark during training.
As such, the transfer type of a method varies with its evaluation benchmark (see Tab.~\ref{tab:dataset-info} for datasets of which categories and images each SLOWP approach has access to during training).

As MaskCLIP and ReCo do not explicitly define background categories, 
we classify the pixels as background if their highest class probability is lower than a certain threshold $t$. 
We set $t$ as 0.25 and 0.9 for MaskCLIP and ReCo (see the supp. mat. for more details on how $t$ is selected).

We evaluate on popular segmentation benchmarks including COCO, CoCA, VOC2012, and large-scale ImageNet-S datasets. Additionally, we also evaluate on the object categories (\eg car, person) in Cityscapes (denoted Cityscapes$_{obj}$).
In addition to SLOWP methods, 
we also compare with state-of-the-art unsupervised semantic segmentation (USS) including MaskContrast~\cite{van2021iccv} and MaskDistill~\cite{gansbeke2022maskdistill} on the VOC2012 and ImageNet-S benchmarks, as they share with SLOWP the similar goal of training without manual annotations.

In Tab.~\ref{tab:comp-slowp}, we compare NamedMask to previous SLOWP methods on COCO, CoCA, and Cityscapes$_{obj}$ benchmarks. We make two observations: (i) ReCo and \methodName, which have access to the category names, outperform MaskCLIP, which is unaware of the concepts of the target benchmarks during training;
(ii) when comparing the two name-only transfer methods, \methodName performs better than ReCo by a large margin on each dataset.

In Tab.~\ref{tab:comp-voc}, 
we report the results of \methodName on the VOC2012 validation split. Our approach shows favourable performance over the existing models for both SLOWP and USS.
In detail, while the previous SLOWP methods fall behind the state-of-the-art USS models, \methodName outperforms them by some ($\approx$13.4 mIoU).
We also observe that the proposed method is competitive on ImageNet-S, which consists of significantly more number of categories than VOC2012. 
Here, \methodName corresponds to the \textit{name-and-image transfer} setting since it has implicit access to unlabelled images from the target distribution through its use of SelfMask (which is bootstrapped from DINO). Similarly, ReCo is categorised as \textit{name-and-image transfer}, as it uses ImageNet1K training images for constructing classifiers.
With the caveat that each method has access to different information, \methodName outperforms the state-of-the-art methods by 15.5, 14.7, and 11.9 mIoU on ImageNet-S$_{50}$ (in Tab.~\ref{tab:comp-imagenet-s50}), ImageNet-S$_{300}$ (in Tab.~\ref{tab:comp-imagenet-s300}), and ImageNet-S$_{919}$ (in Tab.~\ref{tab:comp-imagenet-s919}), respectively.

For qualitative results, we show sample visualisations of our method in Fig.~\ref{fig:teaser}.
More visualisation examples including failure cares are shown in the supplementary material.
\setlength\tabcolsep{2pt}
\begin{table}[!t]
\centering
\begin{tabular}{@{}lcccccc@{}}
\toprule
model & transfer type & mIoU & S & MS & ML & L\\ \midrule
\textit{USS} & &      &      &      &      &      \\
MDC & - & 3.6  & 0.4  & 2.6  & 3.8  & 4.9  \\
MDC$^\dagger$  & - & 14.3 & 2.6  & 10.9 & 14.6 & 19.1 \\
PiCIE & - & 4.5  & 0.2  & 3.1  & 5.0  & 5.3  \\
PiCIE$^\dagger$ & - & 17.6 & 4.4  & 13.1 & 20.1 & 23.1 \\
MaskContrast & - & 24.2 & \textbf{12.2} & 25.6 & 24.7 & 20.4 \\
LUSS-S & - & 29.3 & 6.6  & 25.0 & 33.2 & 32.6 \\
LUSS-P & - & \textbf{32.0} & 9.7  & \textbf{26.2} & \textbf{36.5} & \textbf{40.5} \\ \midrule
\textit{SLOWP}       &      &      &      &      &   \\
MaskCLIP$^*$  & zero-shot & 17.9 & 3.6  & 13.1 & 18.6 & 20.6 \\
ReCo$^*$$^\dagger$ & name-and-image & 22.6 & 10.0 & 24.6 & 22.1 & 18.8 \\
\methodName & name-and-image & \textbf{47.5} & \textbf{23.5} & \textbf{48.7} & \textbf{49.3} & \textbf{38.0} \\ \bottomrule
\end{tabular}
\caption{
Comparison to existing \textit{USS} and \textit{SLOWP} methods on the ImageNet-S$_{50}$ benchmark.
Scores for \textit{USS} are drawn from LUSS~\cite{gao2021arxiv}.
We also report mIoU under different object sizes from small (S), medium-small (MS), medium-large (ML), and large (L).
$^*$Re-implemented and adapted by us to predict a background class.
$^\dagger$initialises the encoder with supervised ImageNet (for MDC and PiCIE) or Stylized-ImageNet pre-training (for ReCo).
Best score for each column within a same method type is in \textbf{bold}.
}
\vspace{-3mm}
\label{tab:comp-imagenet-s50}
\end{table}

\begin{table}[!t]
\centering
\begin{tabular}{@{}lcccccc@{}}
\toprule
model & transfer type & mIoU & S & MS & ML & L\\ \midrule
\textit{USS} &  &      &      &     &      & \\
LUSS-S  & - & 16.0 & 2.8 & 12.0 & 16.4 & 21.7 \\
LUSS-P  & - & \textbf{18.1} & \textbf{4.2} & \textbf{13.6} & \textbf{19.5} & \textbf{23.5} \\ \midrule
\textit{SLOWP} &  &     &      &      &      \\
MaskCLIP$^*$ & zero-shot & 1.6 & 0.4 & 0.6  &  1.2 & 2.5 \\
ReCo$^*$$^\dagger$ & name-and-image & 8.5 & 5.4 & 9.7  & 8.4  & 5.6  \\
\methodName & name-and-image & \textbf{32.8} & \textbf{9.9} & \textbf{29.1} & \textbf{34.9} & \textbf{26.0} \\ \bottomrule
\end{tabular}
\caption{Evaluation on the ImageNet-S$_{300}$ benchmark.
$^*$Re-implemented and adapted by us to predict a background class.
$^\dagger$initialises the encoder with Stylized-ImageNet pre-training.
Best score for each kind of methods is in \textbf{bold}.}
\vspace{-3mm}
\label{tab:comp-imagenet-s300}
\end{table}

\begin{table}[!t]
\centering
\begin{tabular}{@{}lcccccc@{}}
\toprule
model & transfer type & mIoU & S & MS & ML & L\\ \midrule
\textit{USS} &  &      &     &      &      &      \\
LUSS-S & - & 6.6 & 1.3 & 4.6 & 7.1 & 8.4 \\
LUSS-P & - & \textbf{11.0} & \textbf{2.4} & \textbf{8.3} & \textbf{11.9} & \textbf{13.4} \\ \midrule
\textit{SLOWP} & &      &     &      &      &      \\
MaskCLIP$^*$ & zero-shot & 0.5 & 0.1 & 0.2 & 0.3 & 0.8 \\
ReCo$^*$$^\dagger$ & name-and-image & 3.8 & 2.6 & 4.6 & 3.6 & 2.5 \\
\methodName  & name-and-image & \textbf{22.9} & \textbf{5.1} & \textbf{19.4} & \textbf{24.4} & \textbf{19.8}\\ \bottomrule
\end{tabular}
\caption{Evaluation on the ImageNet-S$_{919}$ benchmark.
$^*$Re-implemented and adapted by us to predict a background class.
$^\dagger$initialises the encoder with Stylized-ImageNet pre-training.
\methodName is able to segment reasonably well even when numerous categories (\ie 919 classes) are present in the target dataset. Highest score for each type of methods is highlighted in \textbf{bold}.}
\vspace{-5mm}
\label{tab:comp-imagenet-s919}
\end{table}

\section{Limitations}

We note several limitations of our approach:
(1) We need to train a new segmenter each time we wish to include another category which is not considered in the previous training of \methodName.
Future work could potentially address this by developing a segmenter that directly predicts embeddings in the shared embedding space of CLIP. 
These could subsequently be used for naming predictions beyond the categories seen during training.
(2) While we primarily focus on object semantic segmentation by leveraging an unsupervised saliency detector, it would strengthen our approach to incorporate cues to
segment ``stuff'' categories such as water, sky, etc.

\section{Broader impact}
\methodName distills segmenters from foundation models.
While powerful, these models have been shown to exhibit biases across different racial and religious groups~\cite{bommasani2021opportunities}.
It is therefore likely that \methodName inherits these biases to some degree.
As such \methodName represents a research prototype that is not appropriate for real-world usage without additional consideration of the deployment setting and the design of appropriate mitigation mechanisms.

\methodName aims to achieve semantic segmentation with a methodology that can be scaled up without the prohibitive cost of manually-collected segment annotation.
In doing so, we hope that it will help enable the deployment of semantic segmentation for applications that yield positive societal impact.
As with many powerful computer vision technologies, however, \methodName is a tool that is subject to \textit{dual use} and is therefore vulnerable to abuse.
We are likely unable to anticipate all such possible abuses, but examples could include applications that entail unlawful surveillance.
\section{Conclusion}
In this work, we introduced \methodName, a method for semantic segmentation that is trained by distilling the complementary capabilities of two foundation models, CLIP and DINO, into a single segmenter.
By doing so, \methodName achieves impressive segmentation quality across both single-object and multi-object images without pixel-level annotation.
We demonstrate the effectiveness of \methodName by comparing to prior methods on several benchmarks, where we observe that \methodName achieves a significant boost in segmentation performance.

\newline
\mypara{Acknowledgements.}
This work was performed using resources provided by the Cambridge Service for Data Driven Discovery (CSD3) operated by the University of Cambridge Research Computing Service (www.csd3.cam.ac.uk), provided by Dell EMC and Intel using Tier-2 funding from the Engineering and Physical Sciences Research Council (capital grant EP/T022159/1), and DiRAC funding from the Science and Technology Facilities Council (www.dirac.ac.uk).
GS is supported by AI Factory, Inc. in Korea.
GS would like to thank Guanqi Zhan for proof-reading.
SA would like to acknowledge the support of Z. Novak and N. Novak in enabling his contribution.

\bibliography{refs}
\newpage

\begin{appendices}
In this supplementary material, we describe additional details on our use case of the LAION-5B dataset as an unlabelled image collection (Sec.~\ref{sec:laion}), how we disambiguate homonyms among the ImageNet-S category names (Sec.~\ref{sec:category-names}), and prompt engineering considered to extract textual features for image retrieval (Sec.~\ref{sec:prompt-engineering}). In Sec.~\ref{sec:copy-paste}, we conduct an experiment on the maximum number of images used for copy-paste augmentation and   
we detail how we decide a background threshold for MaskCLIP and ReCo in  Sec.~\ref{sec:background}.
Lastly, we visualise more examples of \methodName including failure cases (Sec.~\ref{sec:visualisations}).

\section{Preprocessing LAION-5B}~\label{sec:laion}
As described in Sec. 4.1 in the paper, we use the LAION-5B dataset~\cite{schuhmann2022web} to build image archives for 919 categories in the ImageNet-S benchmark~\cite{gao2021arxiv} with CLIP. Preparing a subset of LAION-5B containing the ImageNet-S categories consists of two main steps: downloading images from the official LAION search demo, and filtering human images out from the downloaded images. Each step is detailed in the following.

\begin{figure*}[!t]
    \centering
    \includegraphics[width=.98\textwidth]{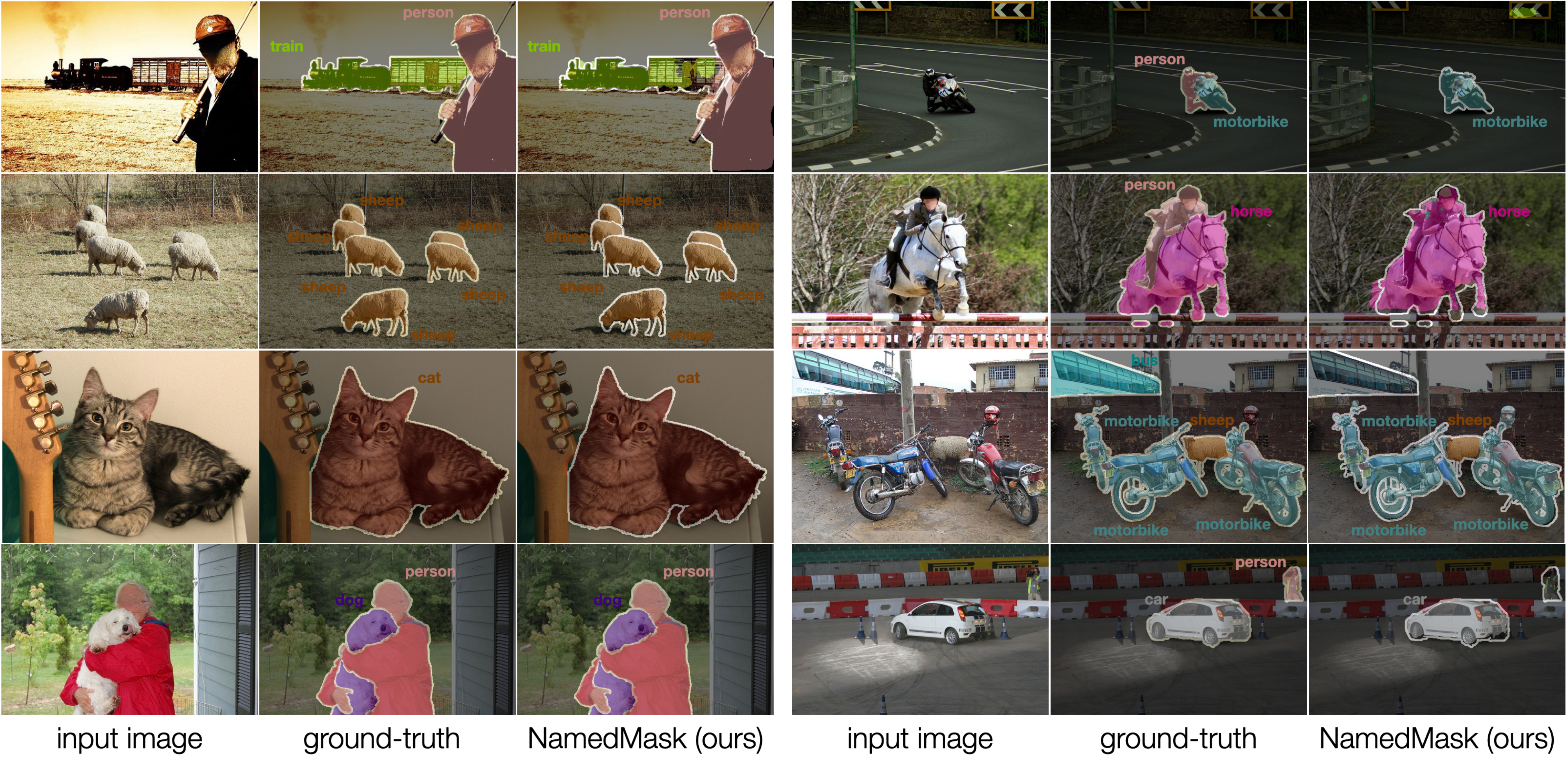}
    \caption{\textbf{Qualitative results of \methodName on VOC2012.} Left: successful cases. Right: typical failure cases. On the ground-truth masks and the predictions, we show a category name of each (predicted) object. Human faces are blurred and ignored regions (\ie object boundaries) are highlighted in white. Best viewed in colour.
    }
    \vspace{-3mm}
    \label{fig:visualisations}
\end{figure*}

\mypara{Downloading images.}
To download LAION-5B images, we use the official demo page of LAION-5B\footnote{\url{https://rom1504.github.io/clip-retrieval}} to download metadata including an image URL.
For this, we search each category name with several search options.
In detail, to ensure that no inappropriate images are retrieved from the LAION-5B database, we use \textit{Safe mode} and \textit{Remove violence}, which filter out NSFW and harmful images, respectively.
We also apply \textit{Hide duplicate urls}, and \textit{Hide (near) duplicate images} functions to diversify retrieved images.
Then, we download top 500 images for each category which have the highest similarity score with a given category text embedding by using their URLs in the metadata. 
We note that there are some URLs which are broken in which cases we simply omit the URL and download the image with the next highest similarity score. 

\mypara{Filtering images with human faces.} To avoid using images containing personal information, we utilise a face detector to filter out human images. For this we use ResNet50-based RetinaFace~\cite{Deng2020CVPR} provided by the open source InsightFace package.\footnote{\url{https://github.com/deepinsight/insightface}} As a result, we obtain 425,618 images in total, with 463 images for the average number of images per category and 326.0 and 360.5 for an average height and width of the images.

\section{Disambiguating ImageNet-S categories}~\label{sec:category-names}
In the main paper, we consider the ImageNet-S benchmark for evaluating our model. We note that there are two pairs of categories sharing a same name with different meanings: \textit{crane} (bird) and \textit{crane} (tower), and \textit{cardigan} (sweater) and \textit{cardigan} (a dog breed).
As this can negatively affect the image retrieval step for constructing image archives, we clarify the meaning of the words by changing \textit{crane} (bird), \textit{crane} (tower), \textit{cardigan} (sweater), and \textit{cardigan} (a dog breed) to \textit{crane bird}, \textit{tower crane}, \textit{cardigan sweater}, and \textit{cardigan welsh corgi} respectively, for collecting images from LAION-5B. 

\section{Prompt engineering}\label{sec:prompt-engineering}
Following~\cite{zhou2022maskclip, shin2022reco}, we use 85 templates to ensemble textual features for a given category, \eg ``{\fontfamily{qcr}\selectfont a photo of a {category}}'' and ``{\fontfamily{qcr}\selectfont a drawing of the {category}}''.
In detail, to produce an ensemble of text features for a category, we average and L2-normalise all of the normalised textual features extracted with the 85 templates. The resulting feature is used for retrieving images by comparing to a normalised visual feature from a CLIP image encoder.

\begin{table}[!t]
\centering
\begin{tabular}{ccc}
\hline
copy-paste        & $n_{max}$ & mIoU \\ \hline
\rowcolor{light_gray}
\xmark            & -     & 56.6 \\
\multirow{5}{*}{\checkmark} & 2     & \textbf{58.7} \\
                  & 4     & 58.3 \\
                  & 6     & 58.5 \\
                  & 8     & 58.5 \\
                  & 10    & 58.5 \\ \hline
\end{tabular}
\caption{The effect of the maximum number of images used for copy-paste on VOC2012. For reference, \methodName trained without copy-paste is highlighted in \color{lightgray}{gray}.}
\vspace{-3mm}
\label{tab:copy-paste-n-images}
\end{table}

\section{Hyperparameter selection for copy-paste}\label{sec:copy-paste}
When we consider copy-paste augmentation for training \methodName, we set a hyperparameter for the maximum number of images $n_{max}$ used for the copy-pate operation. For instance, when $n_{max}$ is set to 10, we randomly select 1 to 10 images to be used for copy-paste at each iteration. It is worth noting that when 1 is selected, it means no copy-paste is applied at the iteration.
To investigate the effect of $n_{max}$ on performance of our model, we train \methodName with different $n_{max}$ values among \{2, 4, 6, 8, 10\} and evaluate on the VOC2012 training split.
For reference, we also report a model which is trained without copy-paste.

As can be noted In Tab.~\ref{tab:copy-paste-n-images}, while the models trained with copy-paste always show superior performance than the one trained without copy-paste, setting $n_{max}$ to 2 allows for the best performance. For this reason we fix $n_{max}$ to 2 for our models in the main paper.

\section{Finding a background threshold}\label{sec:background}
\begin{figure}[!t]
     \centering
     \begin{subfigure}[b]{0.222\textwidth}
     \centering
     \includegraphics[width=\textwidth]{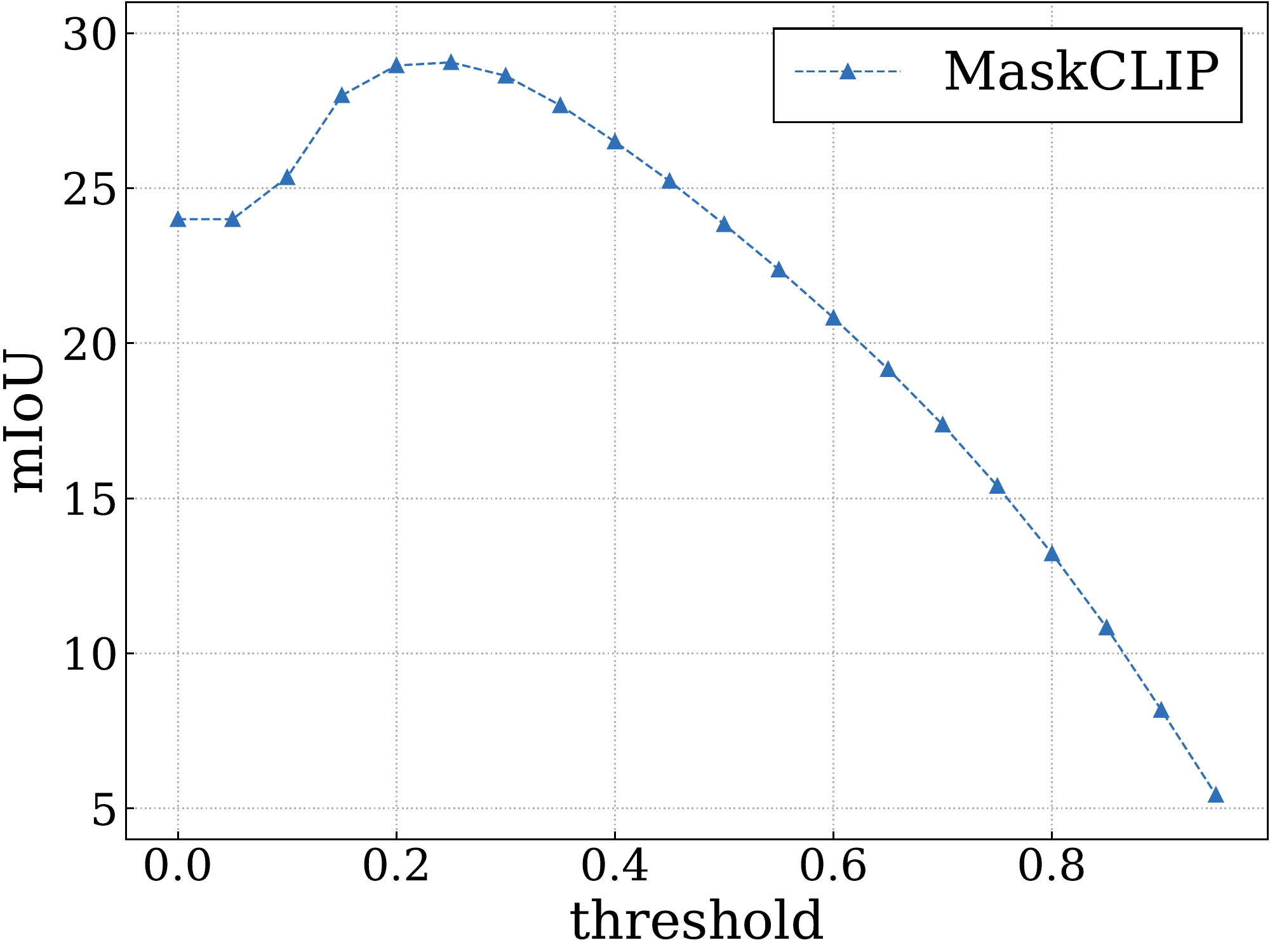}
     \end{subfigure}
     \hspace{5pt}
     \begin{subfigure}[b]{0.222\textwidth}
     \centering
     \includegraphics[width=\textwidth]{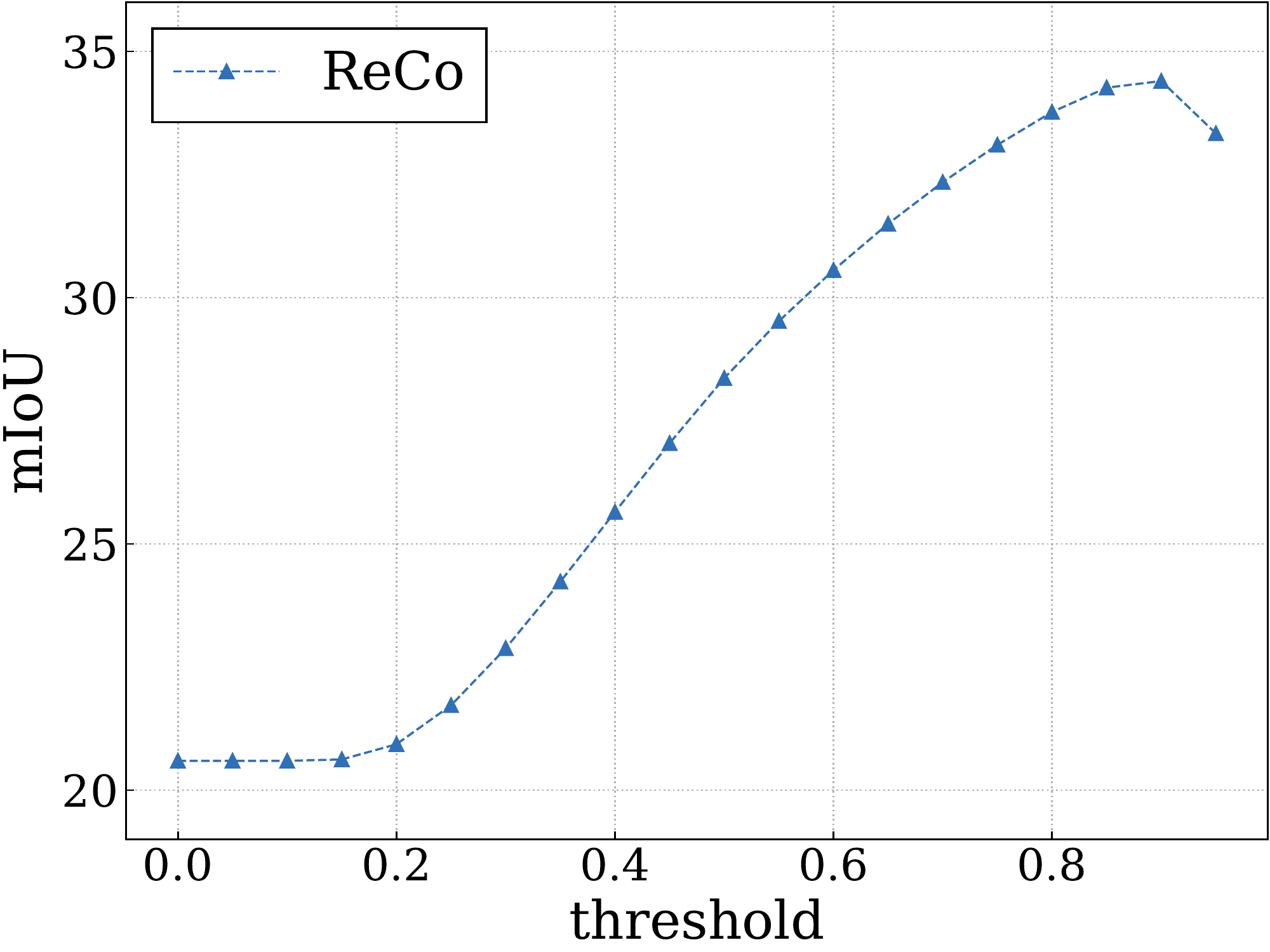}
     \end{subfigure}
     \vspace{-0.5mm}
     \caption{\textbf{Finding an optimal background threshold for MaskCLIP (left) and ReCo (right).} For both cases, we vary a threshold from 0.0 to 0.95 with an interval of 0.05 below which is regarded as background. We observe that performance of both methods varies greatly depending on a threshold.
    }
    \vspace{-3mm}
     \label{fig:background-threshold}
\end{figure}

In Sec. 4.4 of the main paper, we compare our model to MaskCLIP~\cite{zhou2022maskclip} and ReCo~\cite{shin2022reco} on VOC2012 and ImageNet-S, both of which contain a background class.
As MaskCLIP and ReCo do not explicitly predict a background category, we treat pixels whose maximum class probability is lower than a certain threshold as background.
To find the best threshold for each method, we evaluate each model on the VOC2012 training set with a varying threshold $t$ between [0, 0.95] with an interval of 0.05. As shown in Fig.~\ref{fig:background-threshold}, 0.25 and 0.9 allows for the best performance for MaskCLIP and ReCo respectively, we therefore use these thresholds to decide a background pixel throughout our experiments.

\section{More visualisation samples}\label{sec:visualisations}
In Fig.~\ref{fig:visualisations}, we visualise qualitative examples of our model on VOC2012. On the left, we show successful cases whereas on the right, we show typical failure cases.

We note that our model struggles to distinguish a category object and another object of different categories if they tend to co-occur in many cases. For example, a motorbike tends to appear with a person riding it or a horse is inclined to be present with a rider. We conjecture that this is due to our use of a saliency detector for generating (initial) pseudo-masks. That is, pseudo-masks generated by a salient object detector do not separate salient regions based on their semantics, but rather group dominant regions as a whole regardless of their meaning.

To overcome this weakness, it might be helpful to introduce a language-based attention mechanism as considered in~\cite{shin2022reco} for refining pseudo-masks from a saliency model.

\end{appendices}

\end{document}